\documentclass{article}
\usepackage{hyphenat}
\usepackage{microtype}
\usepackage{graphicx}
\usepackage{subcaption}
\usepackage{booktabs}
\usepackage{xcolor, colortbl}
\usepackage{enumitem}
\usepackage{tikz}
\usepackage{amsmath}
\usepackage{amssymb}
\usepackage{bbm}
\usepackage{float}
\usepackage[table]{xcolor}
\usepackage{array}
\usepackage{ragged2e}
\usepackage{tabularx}
\usepackage{multirow}
\usepackage{makecell}

\usepackage{hyperref}

\usepackage[preprint]{icml2026}
\usepackage{amsmath}
\usepackage{amssymb}
\usepackage{mathtools}
\usepackage{amsthm}
\usepackage[capitalize,noabbrev]{cleveref}

\theoremstyle{plain}

\theoremstyle{definition}

\theoremstyle{remark}

\usepackage[textsize=tiny]{todonotes}
\usetikzlibrary{positioning,fit,backgrounds,arrows.meta}

\makeatletter
\g@addto@macro\normalsize{%
  \setlength{\abovedisplayskip}{4pt plus 1pt minus 1pt}%
  \setlength{\belowdisplayskip}{4pt plus 1pt minus 1pt}%
  \setlength{\abovedisplayshortskip}{2pt plus 1pt}%
  \setlength{\belowdisplayshortskip}{2pt plus 1pt}%
}
\makeatother

\icmltitlerunning{MorphStrata: Layer-Specific Perturbations for Generating Morphence Students in Time-Series Moving Target Defense}

\begin{document}

\twocolumn[
  \icmltitle{MorphStrata: Layer-Specific Perturbations for Generating\\Morphence Students in Time-Series Moving Target Defense}

  \icmlkeywords{Moving Target Defense, Adversarial Robustness, Time-Series Forecasting, Transformers, Transformer Models, Machine Learning, Layer-Specific Perturbation, Morphence}

  \begin{icmlauthorlist}
    \icmlauthor{Abhishek Bhardwaj}{sjsu}
    \icmlauthor{Arnav Doshi}{sjsu}
    \icmlauthor{Anusri Nagarajan}{sjsu}
    \icmlauthor{Thanh Quynh Nhu Ta}{sjsu-ce}
    \icmlauthor{Mohammad Masum}{sjsu-ads}
    \icmlauthor{Robert Chun}{sjsu}
    \icmlauthor{Jaydip Sen}{praxis}
    \icmlauthor{Saptarshi Sengupta}{sjsu}
  \end{icmlauthorlist}

  \icmlaffiliation{sjsu}{Department of Computer Science, San Jos\'{e} State University, San Jos\'{e}, CA, USA}
  \icmlaffiliation{sjsu-ce}{Department of Computer Engineering, San Jos\'{e} State University, San Jos\'{e}, CA, USA}
 \icmlaffiliation{sjsu-ads}{Department of Applied Data Science, San Jos\'{e} State University, San Jos\'{e}, CA, USA}
  \icmlaffiliation{praxis}{Praxis Business School, Kolkata, India}

  \icmlcorrespondingauthor{Abhishek Bhardwaj}{abhishek.bhardwaj@sjsu.edu}
  \icmlcorrespondingauthor{Saptarshi Sengupta}{saptarshi.sengupta@sjsu.edu}

\vskip 0.3in

]

\printAffiliationsAndNotice{}

\begin{abstract}
Time-series forecasting models remain vulnerable to gradient-based adversarial attacks, while existing defense mechanisms typically incur a trade-off in robustness for bounded response and compute cost. The problem is pronounced in Moving Target Defense, where maintaining multiple randomized model instances substantially exacerbates the training overhead.
In this work, we introduce MorphStrata, a student generation strategy with selective, layer-specific stochastic noise injection that extends the traditional Morphence defense. MorphStrata uses a Transformer backbone as the teacher and perturbs randomly selected architectural blocks to create structured heterogeneity across student models in response to varied data distributions and threat models.
We evaluate against vanilla Transformer and Morphence backbones on a suite of benchmarks including the Jena Climate (JENA), Electricity Load Diagrams (ECL), and Appliances Energy Prediction (AEP) using FGSM, BIM and PGD attacks across multiple attack strengths.
Across datasets and attack regimes, the proposed ensemble maintains comparable adversarial RMSE. Specifically, for high entropy, periodic datasets as in the case of the AEP data, MorphStrata achieves the lowest RMSE across all attacks and perturbation budgets, improving over the static baseline by up to 24.11\% and 97.97\% under FGSM and BIM respectively at an epsilon value of 0.5 over 30 randomized trials.
Targeting the layers to generate MorphStrata students accounts for less than 1\% increase in train-times over the Morphence MTD baseline for most of the experiments, while accounting for double digit gains in adversarial RMSE reduction. From the experiments, we also observe a positive correlation between higher pairwise L2 distance (among generated students) and overall defense effectiveness.
In summary, MorphStrata maintains adversarial robustness as an MTD defense at marginal cost deltas when compared to existing baselines. The implementation, including all model training and evaluation code, is available at \url{https://github.com/sjsu-micosys-lab/MorphStrata}.
\end{abstract}

\section{Introduction}

Modern time-series forecasting (TSF) increasingly relies on high-capacity Transformer architectures to model complex temporal dependencies \cite{vaswani2017attention,wen2022transformersTSSurvey,nie2022ts64words}. These models remain vulnerable to gradient-based adversarial perturbations; small input changes can produce large forecasting errors \cite{goodfellow2015fgsm,madry2018pgd,govindarajulu2023targetedTSF}. In regression settings like TSF, errors can accumulate across the prediction horizon rather than corrupting a single label, making this threat especially consequential in energy forecasting and industrial monitoring \cite{siddiqui2020benchmarkingTSAdv,liu2022multivariateTSF}.

Moving Target Defense (MTD) reduces attacker reliability by swapping the exposed model dynamically, preventing an adversary from optimizing against a fixed set of parameters \cite{amich2021morphence}. Morphence instantiates this idea for deep networks by generating a pool of student models from a trained base through controlled Gaussian perturbation, then sampling among them at inference. Its student generation, however, is architecture-agnostic. For Transformer-based forecasting this raises a natural question: does the \emph{location} of parameter perturbation matter for robustness, statistical heterogeneity, and cost?

MorphStrata answers this by introducing layer-specific student generation inspired by Morphence's MTD formulation. Rather than perturbing all parameters indiscriminately, MorphStrata targets distinct Transformer components (attention, feed-forward, normalization) through binary masking, creating structured heterogeneity across the student pool. The goal is not to assert uniform superiority; the comparison with the Vanilla Ensemble is deliberately dataset-dependent and attack-dependent. Instead, we treat perturbation location as a design lever that reshapes the robustness-cost trade-off in ways that aggregate perturbation cannot.

\textbf{Contributions.}
\begin{itemize}[topsep=2pt,itemsep=1pt,parsep=0pt,partopsep=0pt,leftmargin=*]
    \item We introduce MorphStrata, a layer-specific MTD framework for Transformer-based time-series forecasting.
    \item We conduct a trade-off focused evaluation covering adversarial RMSE, statistical ensemble heterogeneity (pairwise L2 distance and differential immunity \cite{sengupta2019mtdeep}), and computational overhead across FGSM, BIM, and PGD attacks.
    \item We show that higher pairwise L2 distance among students correlates positively with defense effectiveness, particularly on AEP, where MorphStrata achieves the largest weight diversity gains.
    \item Layer targeting adds less than 1.1\% wall clock overhead over the Vanilla Ensemble across all nine dataset-attack conditions, making MorphStrata a nearly zero-overhead extension for deployments already committed to MTD.
\end{itemize}

\section{Related Work}

Adversarial attacks on TSF have been studied through both untargeted and targeted formulations \cite{govindarajulu2023targetedTSF,siddiqui2020benchmarkingTSAdv,liu2022multivariateTSF,krishan2024phm_adversarial_mts}. FGSM \cite{goodfellow2015fgsm} and its iterative variants BIM \cite{kurakin2017bim} and PGD \cite{madry2018pgd} are the dominant white-box threat models. Defenses span adversarial training \cite{goodfellow2015fgsm,madry2018pgd}, detection \cite{meng2017magnet,zhao2018keybased}, and ensemble-based inference \cite{lakshminarayanan2017deepEnsembles,gal2016dropoutMC,cohen2019randomizedSmoothing}. MTD-based defenses use model switching to degrade attack transferability; Morphence \cite{amich2021morphence} and its extension Morphence 2.0 \cite{awad2022morphence20} are the closest prior work. MTDeep \cite{sengupta2019mtdeep} formalizes differential immunity as a measure of transferability resistance in MTD ensembles; we adapt this metric to the regression setting. Stronger forecasting accuracy does not imply adversarial robustness \cite{cheng2024robustTSF,zhang2025tsfmsurvey}, motivating defense-specific evaluation that goes beyond clean RMSE.

\section{Preliminaries}

\textbf{Forecasting.} Let $\mathbf{x}_{1:T}$ be a multivariate time series with $d$ variables. A Transformer $f_\theta : \mathbb{R}^{T \times d} \to \mathbb{R}^{H \times d'}$ maps a lookback window to a future horizon $H$. Performance is measured by RMSE.

\textbf{Threat model.} A white-box attacker constructs $\mathbf{x}_{\mathrm{adv}} = \mathbf{x} + \boldsymbol{\delta}$ with $\|\boldsymbol{\delta}\|_\infty \le \epsilon$ to maximize forecasting loss. We evaluate FGSM, BIM, and PGD across $\epsilon \in \{0.1, 0.2, 0.3, 0.5\}$.

\textbf{MTD formulation.} A static model exposes one fixed $f_\theta$ at every inference call. MTD instead samples $\hat{\mathbf{y}} = f_{\theta_t}(\mathbf{x}),\ \theta_t \sim p(\theta)$, so the attacker must optimize in expectation over the pool:
\begin{equation}
\boldsymbol{\delta}^{*} = \arg\max_{\|\boldsymbol{\delta}\|_\infty \le \epsilon}\,
\mathbb{E}_{\theta_t \sim p(\theta)}\!\left[\mathcal{L}(f_{\theta_t}(\mathbf{x}+\boldsymbol{\delta}),\mathbf{y})\right].
\end{equation}
\textbf{Morphence baseline.} Given base parameters $\theta$, Morphence generates student $k$ as $\theta^{(k)} = \theta + \boldsymbol{\eta}^{(k)}$, $\boldsymbol{\eta}^{(k)} \sim \mathcal{N}(\mathbf{0}, \sigma^2 \mathbf{I})$, perturbing the full parameter vector without architectural awareness.

\begin{figure*}[t]
\centering
\resizebox{0.98\textwidth}{!}{%
\begin{tikzpicture}[
  >=Stealth,
  every node/.style={font=\small},
  box/.style={
    draw,
    rounded corners=4pt,
    minimum width=3.2cm,
    minimum height=1.0cm,
    align=center,
    inner sep=6pt
  },
  basenode/.style={
    box,
    fill=violet!15,
    draw=violet!60,
    minimum width=3.2cm,
    minimum height=5.0cm
  },
  inpnode/.style={box, fill=cyan!12,   draw=cyan!55!black,   minimum width=3.2cm},
  attnnode/.style={box, fill=teal!15,  draw=teal!60,         minimum width=3.2cm},
  ffnnode/.style={ box, fill=orange!15,draw=orange!55,        minimum width=3.2cm},
  lnnode/.style={  box, fill=yellow!18,draw=yellow!60!orange, minimum width=3.2cm},
  headnode/.style={box, fill=red!12,   draw=red!55,           minimum width=3.2cm},
  pertnode/.style={box, fill=green!12, draw=green!50!black,   minimum width=3.6cm},
  poolnode/.style={box, fill=gray!10,  draw=gray!55,          minimum width=3.4cm},
  advnode/.style={ box, fill=blue!10,  draw=blue!50,          minimum width=3.6cm},
  samplenode/.style={box,fill=blue!14, draw=blue!55,          minimum width=3.6cm},
  ensnode/.style={ box, fill=violet!15,draw=violet!60,        minimum width=4.2cm},
  outnode/.style={ box, fill=green!15, draw=green!60,         minimum width=3.6cm},
  arr/.style={   -{Stealth[length=5pt,width=4pt]}, gray!80, line width=0.9pt},
  thinarr/.style={-{Stealth[length=4pt,width=3pt]}, gray!70, line width=0.6pt}
]


\node[basenode] (base) at (0, 0) {%
  \textbf{Base}\\[3pt]
  \textbf{Transformer}\\[6pt]
  {\scriptsize Clean training}\\
  {\scriptsize on time-series data}
};

\node[inpnode]  (inp)  at (5.4,  2.8)  {\textbf{Input Projection}\\{\scriptsize Input embedding / projection}};
\node[attnnode] (attn) at (5.4,  1.4)  {\textbf{Attention}\\{\scriptsize Q, K, V projections}};
\node[ffnnode]  (ffn)  at (5.4,  0.0)  {\textbf{FFN}\\{\scriptsize Feed-forward weights}};
\node[lnnode]   (ln)   at (5.4, -1.4)  {\textbf{LayerNorm}\\{\scriptsize Scale / shift parameters}};
\node[headnode] (head) at (5.4, -2.8)  {\textbf{Output Head}\\{\scriptsize Forecast regression head}};

\node[pertnode] (pert) at (11.0, 0.8) {%
  \textbf{Masked Gaussian}\\
  \textbf{perturbation}\\[4pt]
  {\scriptsize $\boldsymbol{\eta}^{(i)} \sim \mathcal{N}(\mathbf{0},\sigma^2\mathbf{I})$}\\[2pt]
  {\scriptsize $\theta_i = \theta + \mathbf{m}_{k_i}\odot \boldsymbol{\eta}^{(i)}$}
};

\node[poolnode] (pool) at (15.6, 0.8) {%
  \textbf{MorphStrata}\\[2pt]
  \textbf{student pool}\\[3pt]
  {\scriptsize $\mathcal{F}_{\mathrm{MS}}=\{f_{\theta_i}\}_{i=1}^{N_s}$}
};

\node[advnode] (advtrain) at (20.4, 0.8) {%
  \textbf{Attack-specific}\\[1pt]
  \textbf{student training}\\[3pt]
  {\scriptsize FGSM / BIM / PGD}\\
  {\scriptsize adversarial training}
};

%
%
\def\rowB{-2.8}

\node[samplenode] (sample) at (20.4, \rowB) {%
  \textbf{MTD inference}\\[3pt]
  {\scriptsize Sample $M$ students}\\
  {\scriptsize from trained pool}
};

\node[ensnode] (ens) at (15.7, \rowB) {%
  \textbf{Ensemble prediction}\\[4pt]
  {\scriptsize $\hat{\mathbf{y}}_t=\dfrac{1}{M}\sum_{m=1}^{M} f_{\theta^{(i_m)}}(\mathbf{X}_t)$}
};

\node[outnode] (out) at (11.0, \rowB) {%
  \textbf{Forecast \&}\\[1pt]
  \textbf{evaluation}\\[3pt]
  {\scriptsize RMSE, statistical heterogeneity,}\\
  {\scriptsize stability, resource metrics}
};


\pgfmathsetmacro{\busX}{2.9}
\pgfmathsetmacro{\mergeX}{8.1}

\draw[arr] (base.east) -- (\busX, 0.0);
\draw[gray!60, line width=0.7pt] (\busX, 2.8) -- (\busX, -2.8);

\draw[thinarr] (\busX,  2.8) -- (inp.west);
\draw[thinarr] (\busX,  1.4) -- (attn.west);
\draw[thinarr] (\busX,  0.0) -- (ffn.west);
\draw[thinarr] (\busX, -1.4) -- (ln.west);
\draw[thinarr] (\busX, -2.8) -- (head.west);

\draw[thinarr] (inp.east)  -- (\mergeX,  2.8);
\draw[thinarr] (attn.east) -- (\mergeX,  1.4);
\draw[thinarr] (ffn.east)  -- (\mergeX,  0.0);
\draw[thinarr] (ln.east)   -- (\mergeX, -1.4);
\draw[thinarr] (head.east) -- (\mergeX, -2.8);

\draw[gray!60, line width=0.7pt] (\mergeX, 2.8) -- (\mergeX, -2.8);

\draw[arr] (\mergeX, 0.8) -- (pert.west);
\draw[arr] (pert.east)    -- (pool.west);
\draw[arr] (pool.east)    -- (advtrain.west);

\draw[arr] (advtrain.south) -- (sample.north);

\draw[arr] (sample.west) -- (ens.east);
\draw[arr] (ens.west)    -- (out.east);

\end{tikzpicture}
}
\caption{MorphStrata pipeline. A cleanly trained base Transformer is used to generate student models by applying masked Gaussian perturbations to selected parameter strata, including input projection, attention, feed-forward, LayerNorm, and output-head components. The resulting student pool is adversarially trained under FGSM, BIM, and PGD. During moving target defense (MTD) inference, $M$ students are stochastically sampled from the trained pool, their predictions are averaged, and forecasting performance is evaluated using RMSE, statistical heterogeneity, stability, and resource metrics.}
\label{fig:morphstrata}
\end{figure*}
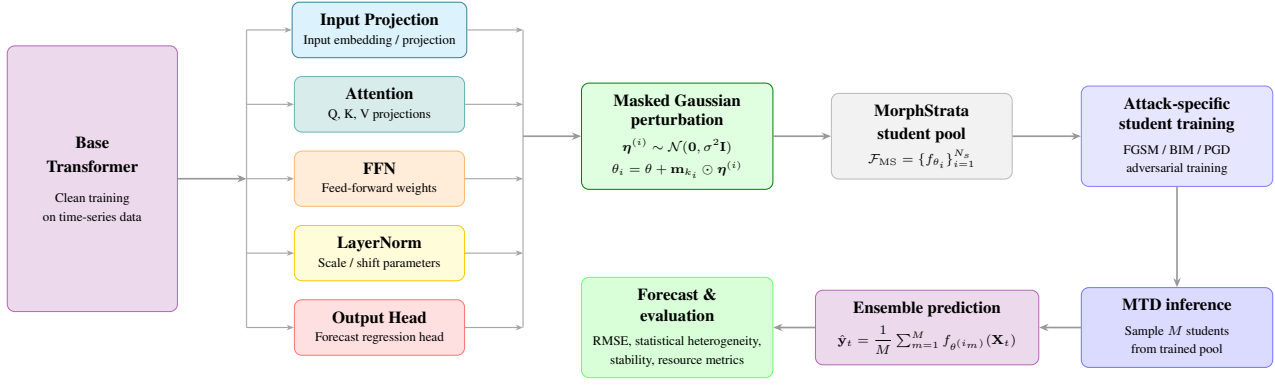

\section{Method}

\subsection{MorphStrata Student Generation}

Let $\{\mathcal{S}_k\}_{k=1}^{K}$ denote parameter strata corresponding to distinct Transformer components: self-attention projections, feed-forward weights, normalization parameters, input projection, and output head. For student $i$, MorphStrata selects stratum $\mathcal{S}_{k_i}$ and applies a binary mask $m_{k_i,j} = \mathbbm{1}[j \in \mathcal{S}_{k_i}]$, giving:
\begin{equation}
\theta^{(i)}_{\mathrm{ms}}
= \theta + \mathbf{m}_{k_i} \odot \boldsymbol{\eta}^{(i)},
\quad \boldsymbol{\eta}^{(i)} \sim \mathcal{N}(\mathbf{0},\sigma^2\mathbf{I}).
\end{equation}
Confining noise to a single functional region produces structurally distinct students; each one is perturbed in a different part of the computational graph. The vanilla baseline applies the same Gaussian noise globally with no mask, so $\theta^{(i)}_{\mathrm{van}} = \theta + \boldsymbol{\eta}^{(i)}$.

\subsection{Adversarial Training and Inference}

Both vanilla and MorphStrata students are adversarially fine-tuned after generation:
\begin{equation}
\min_{\theta_s}\,\mathbb{E}_{(\mathbf{X},\mathbf{y}) \sim \mathcal{D}}\!\left[\mathcal{L}(f_{\theta_s}(\mathbf{X}^{\mathrm{adv}}),\mathbf{y})\right].
\end{equation}
We maintain a fixed pool (no repeated pool regeneration, unlike Morphence), which isolates the effect of generation strategy. At inference, $M$ students are sampled uniformly from the pool and their predictions are averaged:
\begin{equation}
\hat{\mathbf{y}}_t = \frac{1}{M}\sum_{f \in \mathcal{E}_t} f(\mathbf{X}_t),\quad \mathcal{E}_t \subset \mathcal{F},\quad |\mathcal{E}_t|=M.
\end{equation}

\subsection{Algorithmic Summary}

Algorithm~\ref{alg:morphstrata} summarizes MorphStrata student generation and evaluation. The strata $\{\mathcal{S}_k\}_{k=1}^{K}$ correspond to Transformer parameter groups such as attention projections, feed-forward layers, normalization parameters, input projections, and output heads. Students are generated from a common base model, adversarially fine-tuned, and evaluated as a fixed pool without repeated regeneration.

\begin{algorithm}[H]
\footnotesize
\caption{MorphStrata Student Generation and Evaluation}
\label{alg:morphstrata}
\begin{algorithmic}[1]
\REQUIRE Base model $f_{\theta}$, strata $\{\mathcal{S}_k\}_{k=1}^{K}$, noise scale $\sigma$, attacks $\mathcal{A}$, student count $N_s$, ensemble size $M$
\STATE $\mathcal{F}_{0} \leftarrow \emptyset$
\FOR{$i = 1,\dots,N_s$}
    \STATE Select stratum $\mathcal{S}_{k_i}$
    \STATE Sample $\boldsymbol{\eta}^{(i)} \sim \mathcal{N}(\mathbf{0},\sigma^2\mathbf{I})$
    \STATE Define mask $m_{k_i,j}=\mathbbm{1}[j \in \mathcal{S}_{k_i}]$
    \STATE Initialize $\theta_i \leftarrow \theta + \mathbf{m}_{k_i} \odot \boldsymbol{\eta}^{(i)}$
    \STATE $\mathcal{F}_{0} \leftarrow \mathcal{F}_{0} \cup \{f_{\theta_i}\}$
\ENDFOR
\FOR{$a \in \mathcal{A}$}
    \STATE $\mathcal{F}_{a} \leftarrow \emptyset$
    \FOR{each $f_{\theta_i} \in \mathcal{F}_{0}$}
        \STATE Fine-tune $f_{\theta_i}$ using adversarial examples generated by attack $a$
        \STATE Add trained student $f_{\theta_i^{a}}$ to $\mathcal{F}_{a}$
    \ENDFOR
    \FOR{each test input $\mathbf{X}_t$}
        \STATE Sample $\mathcal{E}_t \subset \mathcal{F}_{a}$ uniformly, with $|\mathcal{E}_t|=M$
        \STATE $\hat{\mathbf{y}}_t \leftarrow M^{-1}\sum_{f \in \mathcal{E}_t} f(\mathbf{X}_t)$
    \ENDFOR
    \STATE Compute RMSE, statistical heterogeneity, stability, transferability, and resource metrics for attack $a$
\ENDFOR
\end{algorithmic}
\end{algorithm}

\section{Experiments}

\textbf{Datasets.} We evaluate on three multivariate forecasting benchmarks: Jena Climate (JENA, 60-min weather), Electricity Load Diagrams (ECL, 15-min load), and Appliances Energy Prediction (AEP, 10-min residential energy). All splits are chronological; scaling is fit on the training partition only. Detailed pipelines and temporal structure analysis are in Appendix~\ref{app:dataset-pipelines}.

\textbf{Models.} Three families are compared: a static base Transformer (no defense), the Vanilla Ensemble with global Gaussian perturbation, and MorphStrata. All share the same Transformer architecture and adversarial training procedure; only the student generation strategy differs. Models are implemented in PyTorch.

\textbf{Evaluation.} Each configuration is evaluated over 30 randomized trials. RMSE is the primary metric, reported as mean $\pm$ std. We additionally track pairwise weight L2 distance as a statistical heterogeneity proxy and differential immunity \cite{sengupta2019mtdeep} as a measure of attack non-transferability across the pool. Wall clock time per pipeline stage is measured by per-sample monitoring recording peak CPU RAM, peak VRAM, and stage elapsed time.

\section{Results}

\begin{figure}[!h]
\centering
\includegraphics[width=\columnwidth, keepaspectratio]{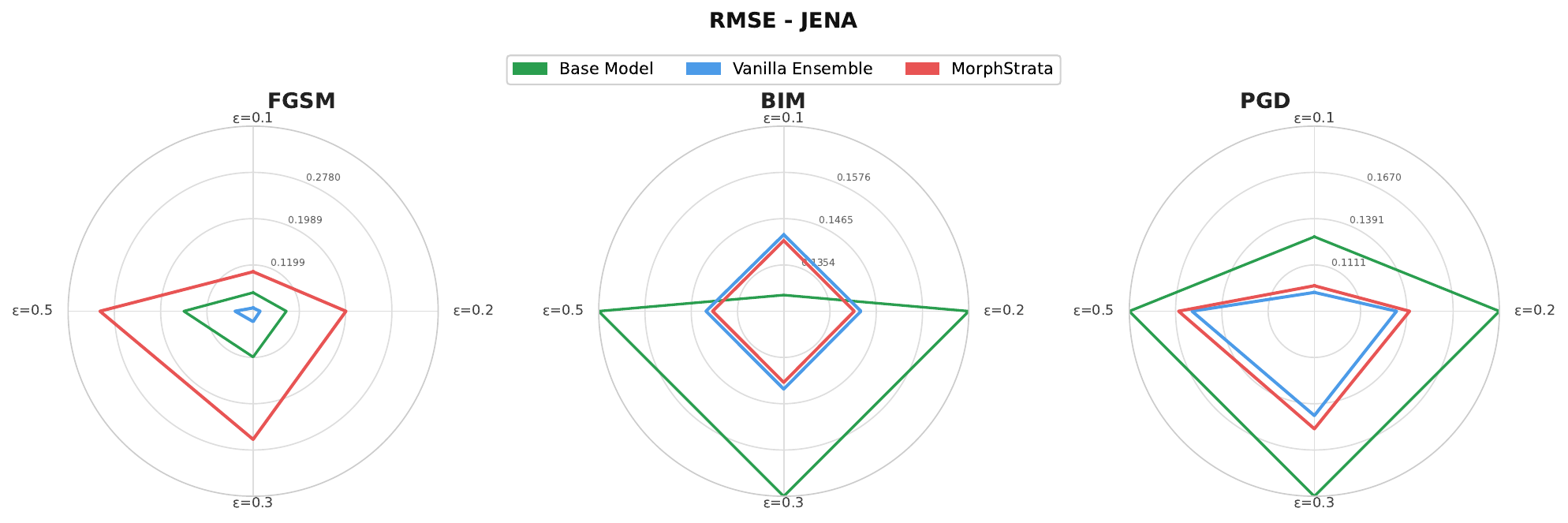}
\caption{RMSE under adversarial attacks on JENA.}
\label{fig:radar_jena}
\end{figure}

\begin{figure}[!h]
\centering
\includegraphics[width=\columnwidth, keepaspectratio]{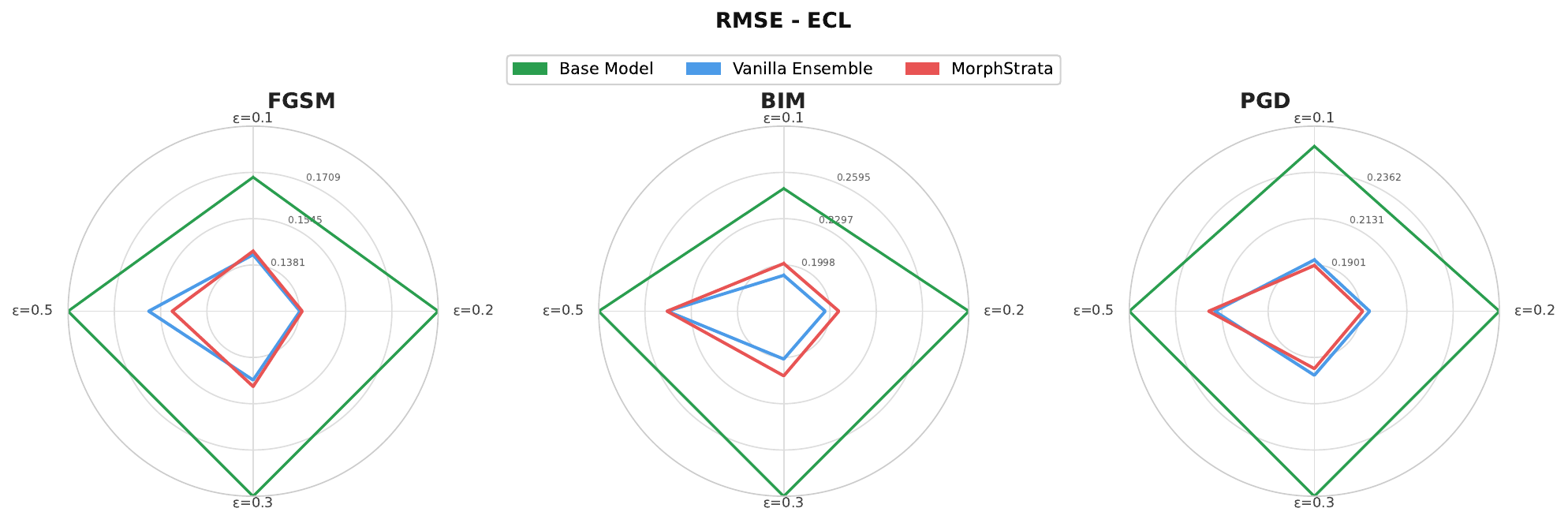}
\caption{RMSE under adversarial attacks on ECL.}
\label{fig:radar_ecl}
\end{figure}

\begin{figure}[!h]
\centering
\includegraphics[width=\columnwidth, keepaspectratio]{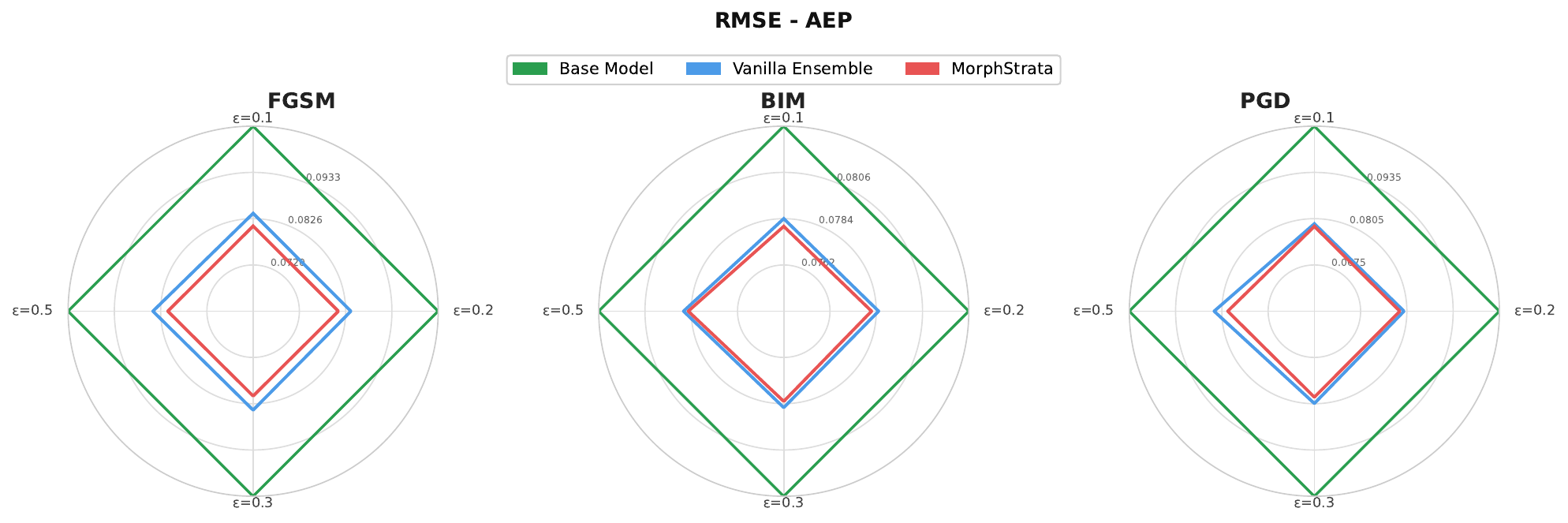}
\caption{RMSE under adversarial attacks on AEP.}
\label{fig:radar_aep}
\end{figure}

\begin{figure}[!h]
\centering
\includegraphics[width=\columnwidth, keepaspectratio]{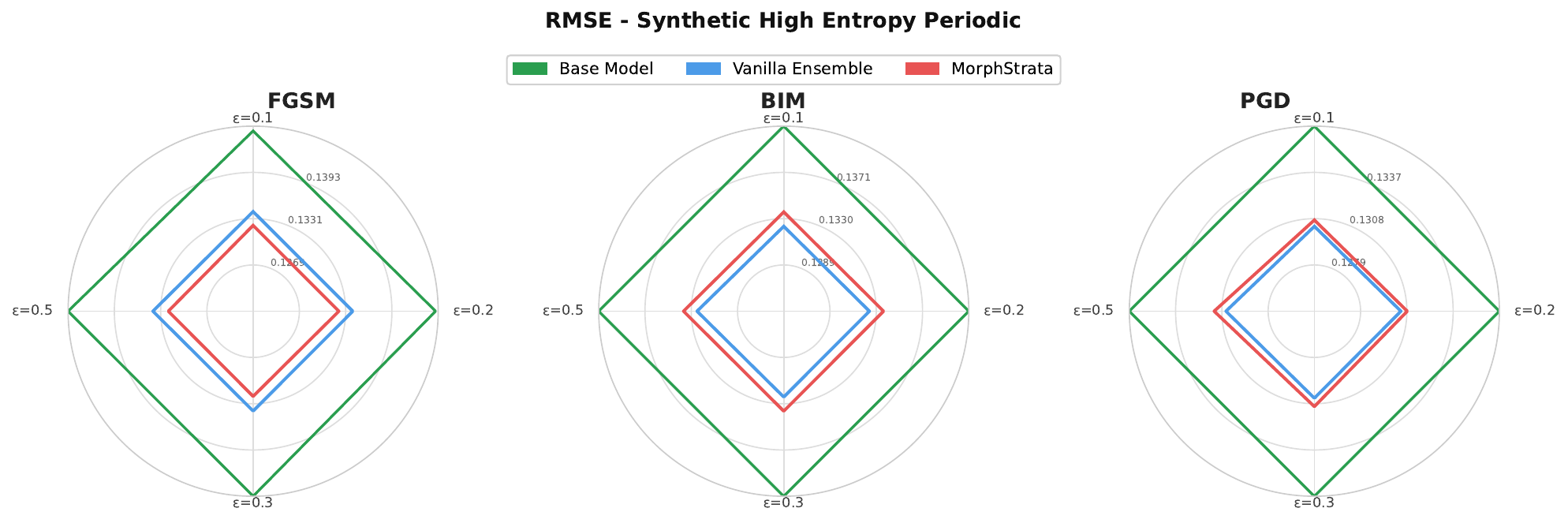}
\caption{RMSE under adversarial attacks on the Synthetic High Entropy Periodic dataset.}
\label{fig:radar_synth_high}
\end{figure}

\begin{figure}[!h]
\centering
\includegraphics[width=\columnwidth, keepaspectratio]{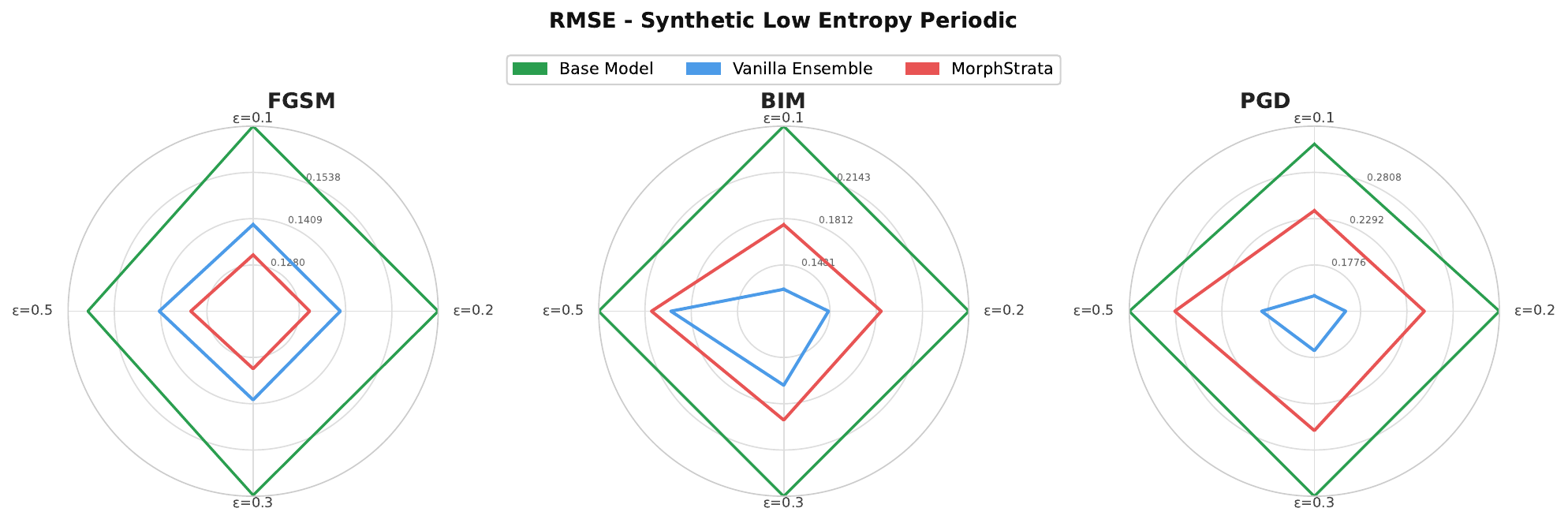}
\caption{RMSE under adversarial attacks on the Synthetic Low Entropy Periodic dataset. Full numerical results for all radar charts provided in Appendix~\ref{app:rmse-tables}.}
\label{fig:radar_synth_low}
\end{figure}

\subsection{Adversarial RMSE}

Figures~\ref{fig:radar_jena}--\ref{fig:radar_synth_low} summarize all dataset-attack
conditions; full RMSE tables with standard deviations are in Appendix~\ref{app:rmse-tables}. Both MTD ensembles substantially reduce adversarial RMSE over the static base model,
particularly under BIM and PGD where the undefended baseline degrades catastrophically
with increasing $\epsilon$. On AEP under BIM at $\epsilon=0.5$, the base model reaches
RMSE 3.865 while MorphStrata holds at 0.079 (a 97.97\% reduction), providing the
strongest empirical evidence that MTD yields substantial returns against iterative threats.

The relative performance of MorphStrata versus the Vanilla Ensemble is dataset-dependent:
the Vanilla Ensemble leads on JENA under FGSM and PGD; the two pipelines are within the
margin of experimental variance on ECL; and MorphStrata achieves the lowest RMSE across
all attacks and perturbation budgets on AEP, with the largest margin under FGSM
(full AEP rows in Appendix~\ref{app:rmse-tables}).

\subsection{Statistical Heterogeneity and Robustness}

The results show a positive correlation between higher pairwise L2 distance among students and defense effectiveness. AEP is the only dataset where MorphStrata consistently produces larger pairwise weight L2 than the Vanilla Ensemble across all three attacks: increases of $+34.2\%$ under FGSM, $+17.9\%$ under BIM, and $+9.0\%$ under PGD, and is also the only dataset where MorphStrata achieves the lowest RMSE in every condition. Full statistical heterogeneity and differential immunity results are in Appendix~\ref{app:diversity}.

\subsection{Temporal Structure and Behavioral Patterns}

The dataset-dependent behavior traces back to temporal structure: JENA has the lowest normalized spectral entropy (0.33) and near-unit lag-1 autocorrelation (0.9996), while AEP has the highest spectral entropy (0.82) and short memory (decay at 260 min). On high-memory, low-entropy datasets like JENA, perturbing a subset of components can disrupt learned long-range dependencies; on AEP, structured layer-specific heterogeneity appears more beneficial. Full temporal analysis, autocorrelation decay curves, and power spectral density plots are in Appendix~\ref{app:temporal-analysis}.

\subsection{Computational Cost}

Layer targeting adds less than 1.1\% wall clock overhead over the Vanilla Ensemble across all nine conditions. MTD itself ranges from $6\times$ to $91\times$ the base training cost, but that cost is shared by both pipelines. For any deployment that has already accepted MTD, MorphStrata incurs negligible cost. Full stage-level breakdowns are in Appendix~\ref{app:cost}.

\section{Conclusion}
MorphStrata introduces layer-specific student generation into the Morphence MTD framework
for Transformer-based time-series forecasting. MTD ensembles substantially outperform the
static base model across all conditions, most strikingly on AEP under iterative attacks
where the base collapses while both ensembles remain stable. MorphStrata's advantage over
the Vanilla Ensemble is conditioned on temporal structure: it consistently achieves lower
RMSE on AEP, is broadly competitive on ECL, and underperforms on JENA under single-step
attacks. Layer targeting adds under 1.1\% training overhead at negligible marginal
inference cost. Future work should examine temporal-structure-aware perturbation scaling,
adaptive student selection, and weight diversity under adaptive attacks.

\section{Impact Statement}
This paper presents work whose goal is to advance the field of Machine Learning. There are many potential societal consequences of our work, none of which we feel must be highlighted here.

\nocite{langley00}
\bibliography{references}
\bibliographystyle{icml2026}


\clearpage
\appendix
\onecolumn

\section{Appendix}
\label{app:appendix}

This appendix contains dataset pipelines, temporal structure analysis, synthetic dataset experiments and generation methodology, full RMSE tables, statistical heterogeneity and differential immunity data, computational cost breakdowns, and memory footprint measurements. The main paper summarizes key findings; all numerical claims cited in the main body are supported here.

\section{Dataset Pipelines}
\label{app:dataset-pipelines}

All datasets are processed chronologically to prevent temporal leakage. Scaling parameters are fit exclusively on the training partition and then applied to validation and test splits. Each dataset is converted into fixed-length sliding-window forecasting samples before training.

\subsection{Jena Climate}

The Jena Climate dataset supports weather forecasting at hourly resolution. The target variable is temperature. Raw measurements are resampled to 60-minute intervals, forward-filled, and backfilled where necessary. Five input features are selected (pressure, temperature, potential temperature, relative humidity, wind speed); the sequence is split 80/20 chronologically, scaled with a MinMax scaler fit on training only, and converted into lookback windows of length 24.

\subsection{Electricity Load Diagrams (ECL)}

ECL captures electricity load at 15-minute resolution for a single meter. The task is multi-step ahead forecasting over a long historical context. Because the input history is long, input patching is applied before Transformer encoding to compress the sequence into a manageable length while preserving coarse temporal structure.

\subsection{Appliances Energy Prediction (AEP)}

AEP captures residential appliance energy usage at 10-minute intervals alongside indoor and outdoor environmental covariates. Two synthetic random variables included in the original dataset are excluded as nuisance features. The target is appliance energy consumption. Separate MinMax scalers are fit for inputs and target on the training split. Input patching is applied as in ECL.

\section{Dataset Temporal Analysis}
\label{app:temporal-analysis}

We characterize each dataset's temporal structure using normalized spectral entropy (Welch PSD, \texttt{nperseg}=512, normalized by $\log(257)$), lag-1 autocorrelation, and autocorrelation decay measured as the first lag where $|\text{ACF}| < 0.1$.

\begin{table}[htbp]
\centering
\caption{Temporal and spectral characteristics. Memory classes: Long ($\geq$1000 min), Medium (200--999 min), Short ($<$200 min). $^\dagger$Synthetic Low (periodic) ACF oscillates due to sinusoidal interference; the 340-min crossing underestimates true memory.}
\label{tab:dataset_spectral}
\begin{tabular}{llcccc}
\toprule
\textbf{Dataset} & \textbf{Group} & \textbf{Norm SE} & \textbf{AC lag-1} & \textbf{Decay (min)} & \textbf{Memory} \\
\midrule
JENA Climate               & Real      & 0.3285 & 0.9996 & 20000+ & Long \\
ECL                        & Real      & 0.8141 & 0.9152 & 30000+ & Long \\
AEP                        & Real      & 0.8152 & 0.7532 & 260    & Medium \\
Syn.\ Low Periodic$^\dagger$ & Synthetic & 0.4295 & 0.9884 & 340    & Medium \\
Syn.\ High Periodic        & Synthetic & 0.9578 & 0.0218 & 10     & Short \\
\bottomrule
\end{tabular}
\end{table}

JENA has the strongest persistence and lowest spectral entropy: a highly structured, periodic signal with very long memory. ECL shares the long-memory property but with higher volatility and higher spectral entropy. AEP decays much faster (260 min) and has the highest spectral entropy among real datasets, reflecting diffuse, locally driven energy consumption patterns. Spectral entropy alone does not separate AEP from ECL (both near 0.81); memory length is the clean differentiator.

This structure helps interpret the RMSE results. On JENA, the dominant long-range temporal pattern is strong enough that confining perturbation to a subset of Transformer components can disrupt learned periodicity; global perturbation is comparatively gentler. On AEP, where no single dominant frequency anchors the signal, layer-specific statistical heterogeneity appears beneficial. ECL, sitting between JENA and AEP on both axes, produces a mixed result with neither pipeline consistently ahead.

\subsection{Autocorrelation Decay}

\begin{figure}[H]
\centering
\includegraphics[width=0.65\textwidth]{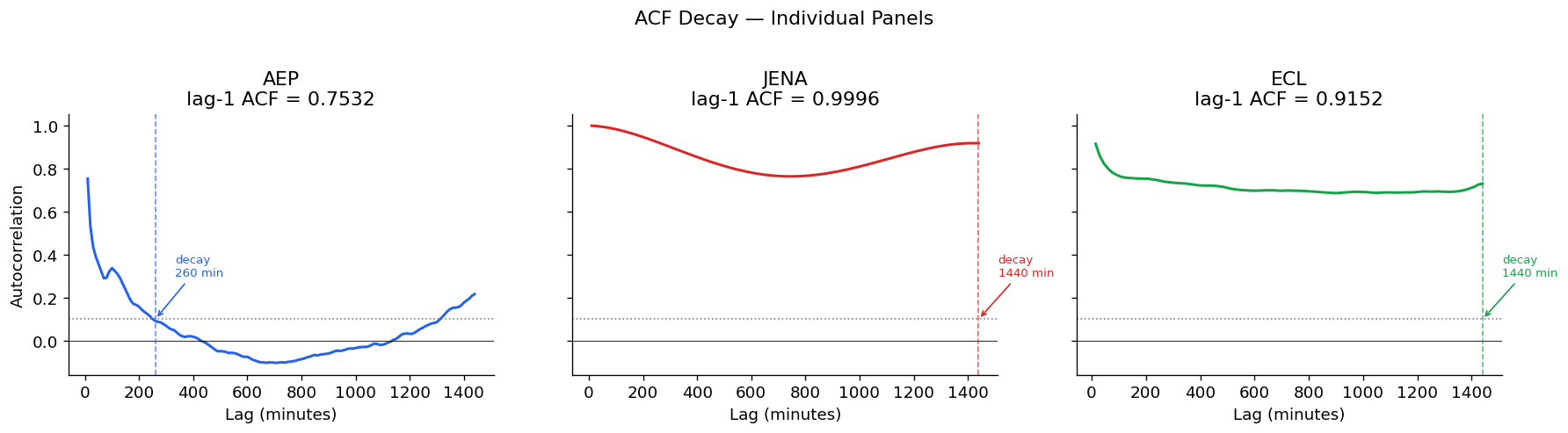}
\caption{Autocorrelation decay for AEP, Jena Climate, and ECL. AEP drops below the 0.1 threshold at 260 minutes; Jena and ECL remain strongly autocorrelated over the full measured window.}
\label{fig:acf-individual}
\end{figure}

\subsection{Power Spectral Density}

\begin{figure}[H]
\centering
\includegraphics[width=0.65\textwidth]{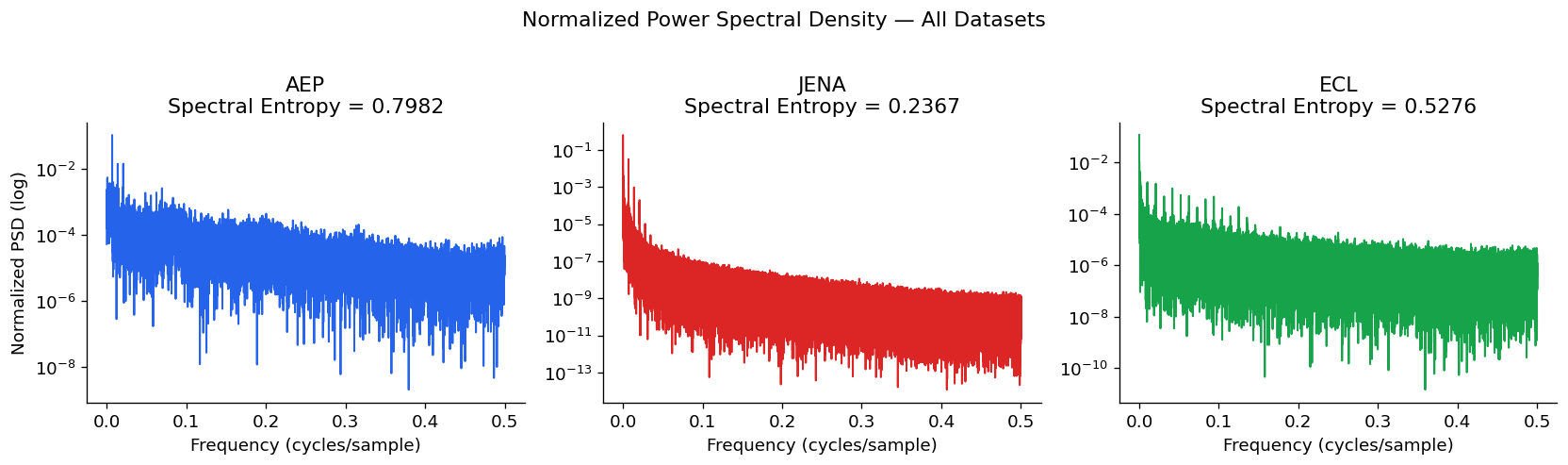}
\caption{Normalized power spectral density for AEP, Jena Climate, and ECL. Spectral entropy is highest for AEP (diffuse) and lowest for Jena (concentrated periodic structure).}
\label{fig:psd-all}
\end{figure}

\subsection{Target Distribution}

\begin{figure}[H]
\centering
\includegraphics[width=0.65\textwidth]{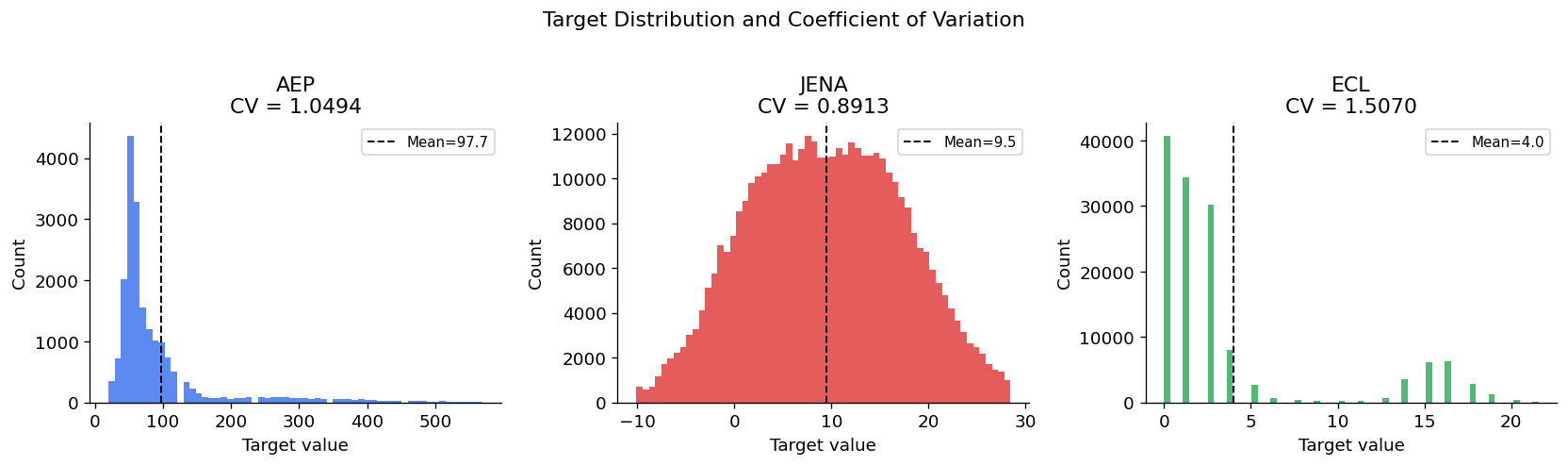}
\caption{Target distribution and coefficient of variation. ECL has the highest normalized volatility; AEP has a right-skewed residential energy distribution; Jena has a smooth unimodal temperature distribution.}
\label{fig:target-distribution}
\end{figure}

\section{Synthetic Dataset Experiments}
\label{app:synthetic}

Two synthetic datasets test whether the behavioral differences observed across JENA, ECL, and AEP generalize to controlled temporal structure variations.

\subsection{Generation Methodology}
\label{app:synthetic-generation}

Both datasets share the following global parameters: $N=19{,}735$ samples (matching AEP), 5 features, 10-minute sampling interval ($\Delta t = 600$\,s), timestamps from 2009-01-01 to 2023-10-14, and random seed 42 (\texttt{numpy.random.default\_rng(42)}).

\textbf{Synthetic-Low (Slow Decay, Periodic).} Each feature $k \in \{0,1,2,3,4\}$ is generated as a function of sample index $t \in \{0, 1, \ldots, N-1\}$:
\begin{equation}
x_k(t) = \underbrace{\sum_{j=1}^{4} A_j \sin\!\left(\frac{2\pi t}{P_j} + \phi_j\right)}_{\text{periodic component}} + \underbrace{\epsilon_k(t)}_{\text{AR(1) noise}} + \underbrace{\tau_k(t)}_{\text{trend}} + 2k
\end{equation}
where $t \in \{0,1,\ldots,N{-}1\}$ is the integer sample index and $P_j$ is expressed in samples. The periodic component uses four dominant sinusoids with fixed periods and amplitudes:
\begin{center}
\footnotesize
\begin{tabular}{cccc}
\toprule
$j$ & Period $P_j$ (samples) & Period (real time) & Amplitude $A_j$ \\
\midrule
1 & 144 & 24 hours & 1.00 \\
2 & 72  & 12 hours & 0.60 \\
3 & 36  & 6 hours  & 0.30 \\
4 & 18  & 3 hours  & 0.15 \\
\bottomrule
\end{tabular}
\end{center}
Phase offsets $\phi_j \sim \text{Uniform}(0, 2\pi)$ are drawn independently per feature. The AR(1) noise uses a slow-decaying process with $\phi=0.97$:
\begin{equation}
\epsilon_k(t) = 0.97 \cdot \epsilon_k(t-1) + \xi_t, \qquad \xi_t \sim \mathcal{N}(0,\, 0.15^2)
\end{equation}
This produces very long autocorrelation memory (lag-10 $\approx 0.78$). Trend: $\tau_k(t) = \text{linspace}(0,\, 0.5(k{+}1),\, N)$; offset: $2k$. Measured properties: spectral entropy (mean across features) $= 2.36$, AC lag-1 $= 0.989$, AC lag-10 $= 0.782$.

\textbf{Synthetic-High (Fast Decay, Diffuse).} Each feature $k$ is generated as:
\begin{equation}
x_k(t) = \underbrace{\sum_{j=1}^{40} a_j \sin\!\left(\frac{2\pi t}{p_j} + \phi_j\right)}_{\text{broadband component}} + \underbrace{\eta_k(t)}_{\text{AR(1) noise}} + \underbrace{w_k(t)}_{\text{random walk}} + 2k
\end{equation}
where $t$ is again the integer sample index. The broadband component uses 40 sinusoids with randomly drawn periods and weak amplitudes spread uniformly across the frequency spectrum:
\begin{equation}
p_j \sim \text{Uniform}(5,\, 9867), \quad a_j \sim \text{Uniform}(0.02,\, 0.15), \quad \phi_j \sim \text{Uniform}(0,\, 2\pi)
\end{equation}
The upper bound $N/2 = 9{,}867$ spans the full Nyquist range. This distributes spectral energy broadly rather than concentrating it at dominant frequencies, which is the defining property of high spectral entropy. AR(1) noise uses fast decay with $\phi=0.35$:
\begin{equation}
\eta_k(t) = 0.35 \cdot \eta_k(t-1) + \zeta_t, \qquad \zeta_t \sim \mathcal{N}(0,\, 1.0^2)
\end{equation}
Random walk: $w_k(t) = \text{cumsum}(\varepsilon)$, $\varepsilon \sim \mathcal{N}(0,\, 0.03^2)$, adding non-stationarity and long-range drift. Offset: $2k$. Measured properties: spectral entropy $= 5.35$, AC lag-1 $= 0.734$, AC lag-10 $= 0.587$.

Spectral entropy was computed using Welch's method (\texttt{nperseg=512}), normalized to a probability distribution before applying Shannon entropy. Both datasets are fully reproducible via the fixed random seed; generation code and CSVs are provided in supplementary materials.

\subsection{Results}

On Synthetic-High and Synthetic-Low, MorphStrata is broadly comparable to vanilla, consistent with the AEP finding that layer-specific heterogeneity benefits diffuse or mixed-entropy signals. Full RMSE tables are in Appendix~\ref{app:rmse-tables}; radar chart comparisons are in Figures~\ref{fig:radar_synth_high} and~\ref{fig:radar_synth_low}.

\section{Model and Training Details}
\label{app:model-training}

All experiments use a shared Transformer architecture: input projection to $d_\text{model}=128$, 4 attention heads, 4 encoder layers, feed-forward dimension 256, pre-norm (norm-first) configuration, dropout 0.1. The same architecture is used for the base model, vanilla students, and MorphStrata students across all three datasets, ensuring the comparison is driven solely by the student generation strategy.

The base model is trained on clean data, and the best checkpoint by validation loss is frozen as the teacher for student generation. Vanilla and MorphStrata students are initialized from this checkpoint and then adversarially fine-tuned. The student pool is fixed after fine-tuning; no repeated pool regeneration is performed.

\section{Attack Implementation}
\label{app:attack-details}

All attacks maximize the MSE forecasting loss with respect to the input under an $\ell_\infty$ budget $\epsilon \in \{0.1,0.2,0.3,0.5\}$. Perturbed inputs are clipped to the valid scaled range $[0,1]$.

\textbf{FGSM.} A randomized single-step variant is used: the input is first perturbed by uniform noise $\mathcal{U}(-\epsilon,\epsilon)$, then one gradient-sign step of size $\alpha=0.02$ is applied.

\textbf{BIM.} Ten projected gradient-sign iterations with step size $\alpha=\epsilon/10$, starting from the clean input. After each step, the perturbation is clipped to the $\epsilon$-ball and the result is clipped to $[0,1]$.

\textbf{PGD.} Identical to BIM but initialized from a random point inside the $\epsilon$-ball.

\section{Full RMSE Tables}
\label{app:rmse-tables}

\begin{table}[h]
\centering
\caption{Jena Climate RMSE (mean $\pm$ std, 30 runs). Bold = best per row.}
\label{tab:jena-rmse}
\footnotesize
\renewcommand{\arraystretch}{1.12}
\setlength{\tabcolsep}{2pt}
\begin{tabular}{|c|c|>{\centering\arraybackslash}p{3.0cm}|>{\centering\arraybackslash}p{3.0cm}|>{\centering\arraybackslash}p{3.0cm}|}
\hline
\rowcolor{gray!20}
\textbf{Attack} & $\boldsymbol{\epsilon}$ & \textbf{Base} & \textbf{Vanilla} & \textbf{MorphStrata} \\
\hline
\multirow{4}{*}{FGSM}
& 0.1 & $0.07265 \pm 0.00023$ & $\mathbf{0.04645 \pm 0.00021}$ & $0.10828 \pm 0.00034$ \\
& 0.2 & $0.09764 \pm 0.00039$ & $\mathbf{0.05272 \pm 0.00019}$ & $0.19944 \pm 0.00051$ \\
& 0.3 & $0.11911 \pm 0.00065$ & $\mathbf{0.05870 \pm 0.00028}$ & $0.25986 \pm 0.00084$ \\
& 0.5 & $0.15934 \pm 0.00085$ & $\mathbf{0.07182 \pm 0.00040}$ & $0.30254 \pm 0.00111$ \\
\hline
\multirow{4}{*}{BIM}
& 0.1 & $0.12817 \pm 0.00022$ & $0.14268 \pm 0.00075$ & $\mathbf{0.14125 \pm 0.00063}$ \\
& 0.2 & $0.21979 \pm 0.00027$ & $0.14284 \pm 0.00064$ & $\mathbf{0.14124 \pm 0.00045}$ \\
& 0.3 & $0.30222 \pm 0.00032$ & $0.14300 \pm 0.00080$ & $\mathbf{0.14140 \pm 0.00058}$ \\
& 0.5 & $0.43441 \pm 0.00061$ & $0.14297 \pm 0.00068$ & $\mathbf{0.14139 \pm 0.00062}$ \\
\hline
\multirow{4}{*}{PGD}
& 0.1 & $0.12817 \pm 0.00022$ & $\mathbf{0.09456 \pm 0.00087}$ & $0.09859 \pm 0.00121$ \\
& 0.2 & $0.21979 \pm 0.00027$ & $\mathbf{0.13286 \pm 0.00116}$ & $0.14070 \pm 0.00088$ \\
& 0.3 & $0.30222 \pm 0.00032$ & $\mathbf{0.14624 \pm 0.00158}$ & $0.15425 \pm 0.00139$ \\
& 0.5 & $0.43441 \pm 0.00061$ & $\mathbf{0.15710 \pm 0.00247}$ & $0.16521 \pm 0.00156$ \\
\hline
\end{tabular}
\end{table}

\begin{table}[h]
\centering
\caption{ECL RMSE (mean $\pm$ std, 30 runs). Bold = best per row.}
\label{tab:ecl-rmse}
\footnotesize
\renewcommand{\arraystretch}{1.12}
\setlength{\tabcolsep}{2pt}
\begin{tabular}{|c|c|>{\centering\arraybackslash}p{3.0cm}|>{\centering\arraybackslash}p{3.0cm}|>{\centering\arraybackslash}p{3.0cm}|}
\hline
\rowcolor{gray!20}
\textbf{Attack} & $\boldsymbol{\epsilon}$ & \textbf{Base} & \textbf{Vanilla} & \textbf{MorphStrata} \\
\hline
\multirow{4}{*}{FGSM}
& 0.1 & $0.16916 \pm 0.00046$ & $\mathbf{0.14172 \pm 0.00096}$ & $0.14299 \pm 0.00097$ \\
& 0.2 & $0.23045 \pm 0.00085$ & $\mathbf{0.13827 \pm 0.00091}$ & $0.13907 \pm 0.00092$ \\
& 0.3 & $0.27207 \pm 0.00068$ & $\mathbf{0.14609 \pm 0.00121}$ & $0.14837 \pm 0.00123$ \\
& 0.5 & $0.29805 \pm 0.00072$ & $0.15870 \pm 0.00128$ & $\mathbf{0.15040 \pm 0.00141}$ \\
\hline
\multirow{4}{*}{BIM}
& 0.1 & $0.24913 \pm 0.00055$ & $\mathbf{0.19305 \pm 0.00072}$ & $0.20067 \pm 0.00045$ \\
& 0.2 & $0.33875 \pm 0.00046$ & $\mathbf{0.19668 \pm 0.00066}$ & $0.20547 \pm 0.00051$ \\
& 0.3 & $0.36828 \pm 0.00043$ & $\mathbf{0.20080 \pm 0.00060}$ & $0.21164 \pm 0.00063$ \\
& 0.5 & $0.39611 \pm 0.00037$ & $\mathbf{0.24511 \pm 0.00116}$ & $0.24527 \pm 0.00133$ \\
\hline
\multirow{4}{*}{PGD}
& 0.1 & $0.24921 \pm 0.00053$ & $0.19259 \pm 0.00051$ & $\mathbf{0.18979 \pm 0.00049}$ \\
& 0.2 & $0.33865 \pm 0.00045$ & $0.19457 \pm 0.00057$ & $\mathbf{0.19113 \pm 0.00072}$ \\
& 0.3 & $0.36821 \pm 0.00035$ & $0.19894 \pm 0.00049$ & $\mathbf{0.19563 \pm 0.00073}$ \\
& 0.5 & $0.39596 \pm 0.00045$ & $\mathbf{0.21664 \pm 0.00069}$ & $0.21967 \pm 0.00096$ \\
\hline
\end{tabular}
\end{table}

\begin{table}[h]
\centering
\caption{AEP RMSE (mean $\pm$ std; 30 runs). Bold = best per row.}
\label{tab:aep-rmse}
\footnotesize
\renewcommand{\arraystretch}{1.12}
\setlength{\tabcolsep}{2pt}
\begin{tabular}{|c|c|>{\centering\arraybackslash}p{3.0cm}|>{\centering\arraybackslash}p{3.0cm}|>{\centering\arraybackslash}p{3.0cm}|}
\hline
\rowcolor{gray!20}
\textbf{Attack} & $\boldsymbol{\epsilon}$ & \textbf{Base} & \textbf{Vanilla} & \textbf{MorphStrata} \\
\hline
\multirow{4}{*}{FGSM}
& 0.1 & $0.10398 \pm 0.00003$ & $0.08386 \pm 0.00019$ & $\mathbf{0.08093 \pm 0.00039}$ \\
& 0.2 & $0.10430 \pm 0.00004$ & $0.08384 \pm 0.00005$ & $\mathbf{0.08095 \pm 0.00024}$ \\
& 0.3 & $0.10491 \pm 0.00005$ & $0.08410 \pm 0.00012$ & $\mathbf{0.08085 \pm 0.00034}$ \\
& 0.5 & $0.10670 \pm 0.00010$ & $0.08441 \pm 0.00007$ & $\mathbf{0.08097 \pm 0.00028}$ \\
\hline
\multirow{4}{*}{BIM}
& 0.1 & $0.30552 \pm 0.00023$ & $0.07841 \pm 0.00008$ & $\mathbf{0.07805 \pm 0.00004}$ \\
& 0.2 & $1.03908 \pm 0.00117$ & $0.07852 \pm 0.00006$ & $\mathbf{0.07819 \pm 0.00006}$ \\
& 0.3 & $2.05804 \pm 0.00241$ & $0.07859 \pm 0.00008$ & $\mathbf{0.07830 \pm 0.00006}$ \\
& 0.5 & $3.86461 \pm 0.00700$ & $0.07877 \pm 0.00009$ & $\mathbf{0.07855 \pm 0.00012}$ \\
\hline
\multirow{4}{*}{PGD}
& 0.1 & $0.30552 \pm 0.00023$ & $0.07909 \pm 0.00017$ & $\mathbf{0.07835 \pm 0.00004}$ \\
& 0.2 & $1.03908 \pm 0.00118$ & $0.07971 \pm 0.00022$ & $\mathbf{0.07854 \pm 0.00005}$ \\
& 0.3 & $2.05807 \pm 0.00241$ & $0.08044 \pm 0.00043$ & $\mathbf{0.07868 \pm 0.00005}$ \\
& 0.5 & $3.86453 \pm 0.00692$ & $0.08268 \pm 0.00089$ & $\mathbf{0.07887 \pm 0.00006}$ \\
\hline
\end{tabular}
\end{table}

\begin{table}[h]
\centering
\caption{Synthetic High Entropy Periodic RMSE (mean $\pm$ std, 30 runs). Bold = best per row.}
\label{tab:synth-high-rmse}
\footnotesize
\renewcommand{\arraystretch}{1.12}
\setlength{\tabcolsep}{2pt}
\begin{tabular}{|c|c|>{\centering\arraybackslash}p{3.0cm}|>{\centering\arraybackslash}p{3.0cm}|>{\centering\arraybackslash}p{3.0cm}|}
\hline
\rowcolor{gray!20}
\textbf{Attack} & $\boldsymbol{\epsilon}$ & \textbf{Base} & \textbf{Vanilla} & \textbf{MorphStrata} \\
\hline
\multirow{4}{*}{FGSM}
& 0.1 & $0.14495 \pm 0.00002$ & $0.13405 \pm 0.00030$ & $\mathbf{0.13219 \pm 0.00011}$ \\
& 0.2 & $0.14525 \pm 0.00004$ & $0.13406 \pm 0.00026$ & $\mathbf{0.13219 \pm 0.00009}$ \\
& 0.3 & $0.14559 \pm 0.00005$ & $0.13409 \pm 0.00023$ & $\mathbf{0.13210 \pm 0.00009}$ \\
& 0.5 & $0.14565 \pm 0.00007$ & $0.13414 \pm 0.00033$ & $\mathbf{0.13206 \pm 0.00011}$ \\
\hline
\multirow{4}{*}{BIM}
& 0.1 & $0.24846 \pm 0.00007$ & $\mathbf{0.13230 \pm 0.00014}$ & $0.13356 \pm 0.00031$ \\
& 0.2 & $0.42712 \pm 0.00026$ & $\mathbf{0.13237 \pm 0.00015}$ & $0.13362 \pm 0.00028$ \\
& 0.3 & $0.59956 \pm 0.00049$ & $\mathbf{0.13239 \pm 0.00016}$ & $0.13363 \pm 0.00025$ \\
& 0.5 & $0.88182 \pm 0.00073$ & $\mathbf{0.13246 \pm 0.00020}$ & $0.13366 \pm 0.00036$ \\
\hline
\multirow{4}{*}{PGD}
& 0.1 & $0.24846 \pm 0.00007$ & $\mathbf{0.13031 \pm 0.00014}$ & $0.13071 \pm 0.00016$ \\
& 0.2 & $0.42712 \pm 0.00026$ & $\mathbf{0.13042 \pm 0.00014}$ & $0.13082 \pm 0.00017$ \\
& 0.3 & $0.59956 \pm 0.00049$ & $\mathbf{0.13045 \pm 0.00012}$ & $0.13098 \pm 0.00016$ \\
& 0.5 & $0.88182 \pm 0.00073$ & $\mathbf{0.13054 \pm 0.00016}$ & $0.13128 \pm 0.00027$ \\
\hline
\end{tabular}
\end{table}

\begin{table}[!h]
\centering
\caption{Synthetic Low Entropy Periodic RMSE (mean $\pm$ std, 30 runs). Bold = best per row.}
\label{tab:synth-low-rmse}
\footnotesize
\renewcommand{\arraystretch}{1.12}
\setlength{\tabcolsep}{2pt}
\begin{tabular}{|c|c|>{\centering\arraybackslash}p{3.0cm}|>{\centering\arraybackslash}p{3.0cm}|>{\centering\arraybackslash}p{3.0cm}|}
\hline
\rowcolor{gray!20}
\textbf{Attack} & $\boldsymbol{\epsilon}$ & \textbf{Base} & \textbf{Vanilla} & \textbf{MorphStrata} \\
\hline
\multirow{4}{*}{FGSM}
& 0.1 & $0.16885 \pm 0.00006$ & $0.13929 \pm 0.00070$ & $\mathbf{0.13079 \pm 0.00092}$ \\
& 0.2 & $0.16804 \pm 0.00012$ & $0.13942 \pm 0.00055$ & $\mathbf{0.13087 \pm 0.00085}$ \\
& 0.3 & $0.16640 \pm 0.00024$ & $0.13982 \pm 0.00079$ & $\mathbf{0.13114 \pm 0.00093}$ \\
& 0.5 & $0.16121 \pm 0.00023$ & $0.14127 \pm 0.00053$ & $\mathbf{0.13248 \pm 0.00082}$ \\
\hline
\multirow{4}{*}{BIM}
& 0.1 & $0.31245 \pm 0.00009$ & $\mathbf{0.13074 \pm 0.00115}$ & $0.17690 \pm 0.00283$ \\
& 0.2 & $0.43343 \pm 0.00015$ & $\mathbf{0.14705 \pm 0.00100}$ & $0.18475 \pm 0.00310$ \\
& 0.3 & $0.52030 \pm 0.00030$ & $\mathbf{0.16800 \pm 0.00105}$ & $0.19289 \pm 0.00393$ \\
& 0.5 & $0.63804 \pm 0.00044$ & $\mathbf{0.19577 \pm 0.00079}$ & $0.20963 \pm 0.00372$ \\
\hline
\multirow{4}{*}{PGD}
& 0.1 & $0.31245 \pm 0.00009$ & $\mathbf{0.14318 \pm 0.00132}$ & $0.23813 \pm 0.00884$ \\
& 0.2 & $0.43343 \pm 0.00015$ & $\mathbf{0.16091 \pm 0.00136}$ & $0.24864 \pm 0.01054$ \\
& 0.3 & $0.52030 \pm 0.00030$ & $\mathbf{0.17005 \pm 0.00130}$ & $0.25912 \pm 0.01116$ \\
& 0.5 & $0.63804 \pm 0.00044$ & $\mathbf{0.18485 \pm 0.00179}$ & $0.28165 \pm 0.01479$ \\
\hline
\end{tabular}
\end{table}

\newpage
\section{Statistical Heterogeneity and Differential Immunity}
\label{app:diversity}

We report two metrics for statistical heterogeneity. Pairwise weight L2 distance is the Euclidean norm between flattened parameter vectors for each pair of students in the pool ):
\begin{equation}
L_2(\theta_i,\theta_j) = \left(\sum_\ell (\theta_{i,\ell}-\theta_{j,\ell})^2\right)^{1/2}.
\end{equation}
Higher L2 indicates that students occupy more separated regions of parameter space, which tends to reduce attack transferability across the pool.

Differential immunity $\delta$ is adapted from MTDeep \cite{sengupta2019mtdeep}, originally proposed for classification. For a fixed attacker $u$ and budget $\epsilon$, we define:
\begin{equation}
\delta(u,\epsilon) = \frac{\max_n \mathrm{RMSE}(n,u) - \min_n \mathrm{RMSE}(n,u)}{\max_n \mathrm{RMSE}(n,u)},
\end{equation}
where $n$ indexes defender students. A high $\delta$ means the pool's best defender performs much better than its worst defender against that attacker, i.e., the attack does not transfer uniformly. We report worst-case $\delta$ across all attackers and all $\epsilon$ values per cell. The adaptation to regression is direct: RMSE replaces classification error, and the ratio preserves the same scale-free interpretation as the original formulation.

\begin{table}[t]
\centering
\caption{Pairwise weight L2 distance and worst-case differential immunity $\delta$. Bold indicates MorphStrata improves over Vanilla. $\Delta L_2$ and $\Delta\delta$ are absolute changes (MorphStrata minus Vanilla).}
\label{tab:diversity}
\footnotesize
\renewcommand{\arraystretch}{1.12}
\setlength{\tabcolsep}{2pt}
\begin{tabular}{|c|c|>{\centering\arraybackslash}m{1.4cm}|>{\centering\arraybackslash}m{1.9cm}|>{\centering\arraybackslash}m{1.4cm}|>{\centering\arraybackslash}m{1.9cm}|>{\centering\arraybackslash}m{2.2cm}|>{\centering\arraybackslash}m{1.4cm}|}
\hline
\rowcolor{gray!20}
\textbf{Dataset} & \textbf{Attack} & \multicolumn{2}{c|}{\textbf{Weight L2}} & \multicolumn{2}{c|}{\textbf{$\delta$ worst}} & $\Delta L_2$ & $\Delta\delta$ \\
\hline
\rowcolor{gray!20}
& & \textbf{Vanilla} & \textbf{MorphStrata} & \textbf{Vanilla} & \textbf{MorphStrata} & & \\
\hline
\multirow{3}{*}{JENA}
& FGSM & 27.24          & 26.47          & 0.0041          & \textbf{0.0397} & $-0.77$ ($-2.8\%$)   & $+0.0356$ \\
\cline{2-8}
& BIM  & 12.58          & \textbf{13.53} & 0.0305          & \textbf{0.0819} & $+0.95$ ($+7.5\%$)   & $+0.0514$ \\
\cline{2-8}
& PGD  & \textbf{73.10} & 73.68          & \textbf{0.1274} & 0.0525          & $+0.58$ ($+0.8\%$)   & $-0.0750$ \\
\hline
\multirow{3}{*}{ECL}
& FGSM & 16.01          & 14.81          & 0.0158          & \textbf{0.1497} & $-1.20$ ($-7.5\%$)   & $+0.1339$ \\
\cline{2-8}
& BIM  & 9.46           & 8.40           & \textbf{0.1418} & 0.0834          & $-1.06$ ($-11.2\%$)  & $-0.0584$ \\
\cline{2-8}
& PGD  & 9.01           & 8.96           & \textbf{0.0635} & 0.0216          & $-0.05$ ($-0.6\%$)   & $-0.0418$ \\
\hline
\multirow{3}{*}{AEP}
& FGSM & 30.48          & \textbf{40.89} & 0.0133          & \textbf{0.0180} & $+10.41$ ($+34.2\%$) & $+0.0047$ \\
\cline{2-8}
& BIM  & 38.63          & \textbf{45.53} & \textbf{0.0384} & 0.0085          & $+6.91$ ($+17.9\%$)  & $-0.0299$ \\
\cline{2-8}
& PGD  & 37.01          & \textbf{40.35} & 0.0090          & \textbf{0.0216} & $+3.34$ ($+9.0\%$)   & $+0.0126$ \\
\hline
\end{tabular}
\end{table}

Several patterns are worth noting. AEP is the only dataset where MorphStrata consistently produces larger pairwise L2 than Vanilla across all three attacks; it is also the only dataset where MorphStrata achieves the lowest RMSE in every condition. This alignment supports the observed positive L2-robustness correlation. On JENA and ECL, L2 changes are small in absolute terms and bidirectional; MorphStrata's RMSE behavior is correspondingly mixed on these datasets.

Differential immunity is less monotone. MorphStrata improves $\delta$ in 5 of 9 cells, with the largest gains on JENA-FGSM and ECL-FGSM, both single-step attack conditions. The JENA-PGD cell is the clearest anomaly: the Vanilla Ensemble already produces pairwise L2 of 73.10, the largest in the entire table, leaving no room for MorphStrata to widen the pool further. Accordingly, the Vanilla Ensemble beats MorphStrata on JENA-PGD in the RMSE tables.

\section{Computational Cost}
\label{app:cost}

Resource metrics are produced by a per-stage monitoring system running on all three dataset pipelines. It samples every 2 seconds and records peak CPU RAM, peak VRAM, and stage wall-clock time. Per-sample inference latency is not included in the monitoring stack. All experiments were conducted on an NVIDIA L4 GPU (Ada Lovelace, 22.5\,GB VRAM) via Google Colab.

\subsection{Layer Targeting Overhead over the Vanilla Ensemble}

The extra cost of MorphStrata over the Vanilla Ensemble is computed as:
\begin{equation}
\text{Extra cost} = \frac{\text{MorphStrata total} - \text{Vanilla total}}{\text{Vanilla total}} \times 100\%.
\end{equation}

\begin{table}[t]
\centering
\caption{Wall-clock cost comparison: Vanilla Ensemble vs.\ MorphStrata. Times are total pipeline duration in minutes; Extra cost is the percentage overhead of MorphStrata over Vanilla.}
\label{tab:wallclock}
\footnotesize
\renewcommand{\arraystretch}{1.12}
\setlength{\tabcolsep}{2pt}
\begin{tabular}{|c|c|>{\centering\arraybackslash}m{2.6cm}|>{\centering\arraybackslash}m{2.6cm}|>{\centering\arraybackslash}m{2.2cm}|}
\hline
\rowcolor{gray!20}
\textbf{Dataset} & \textbf{Attack} & \textbf{Vanilla (minutes)} & \textbf{MorphStrata (minutes)} & \textbf{Extra cost} \\
\hline
\multirow{3}{*}{JENA}
& FGSM & 9.54   & 9.64   & $+1.05\%$ \\
\cline{2-5}
& BIM  & 41.48  & 41.59  & $+0.27\%$ \\
\cline{2-5}
& PGD  & 41.45  & 41.62  & $+0.41\%$ \\
\hline
\multirow{3}{*}{ECL}
& FGSM & 29.20  & 29.27  & $+0.24\%$ \\
\cline{2-5}
& BIM  & 148.43 & 148.85 & $+0.28\%$ \\
\cline{2-5}
& PGD  & 148.34 & 149.08 & $+0.50\%$ \\
\hline
\multirow{3}{*}{AEP}
& FGSM & 29.49  & 29.55  & $+0.20\%$ \\
\cline{2-5}
& BIM  & 159.36 & 159.47 & $+0.07\%$ \\
\cline{2-5}
& PGD  & 159.36 & 159.44 & $+0.05\%$ \\
\hline
\end{tabular}
\end{table}

Across all nine conditions, the extra cost of layer targeting stays under 1.1\%. The masking operation during student initialization is the only additional computation; adversarial training and inference are identical between the two pipelines.

\subsection{MTD Overhead over the Base Model}

For deployments deciding whether to adopt MTD at all, the relevant comparison is the full MorphStrata pipeline against a single base Transformer trained on clean data:
\begin{equation}
\text{MTD overhead} = \frac{\text{MorphStrata total}}{\text{Base training time}}.
\end{equation}

\begin{table}[h]
\centering
\caption{MorphStrata pipeline overhead relative to undefended base training. Vanilla/Base and MorphStrata/Base are multiplicative overheads; MorphStrata/Vanilla is the marginal overhead of MorphStrata over the Vanilla Ensemble.}
\label{tab:mtd-overhead}
\footnotesize
\renewcommand{\arraystretch}{1.12}
\setlength{\tabcolsep}{2pt}
\begin{tabular}{|c|c|>{\centering\arraybackslash}m{1.4cm}|>{\centering\arraybackslash}m{1.6cm}|>{\centering\arraybackslash}m{1.8cm}|>{\centering\arraybackslash}m{1.5cm}|>{\centering\arraybackslash}m{1.9cm}|>{\centering\arraybackslash}m{1.9cm}|}
\hline
\rowcolor{gray!20}
\textbf{Dataset} & \textbf{Attack} & \textbf{Base (minutes)} & \textbf{Vanilla (minutes)} & \textbf{MorphStrata (minutes)} & \textbf{Vanilla / Base} & \textbf{MorphStrata / Base} & \textbf{MorphStrata / Vanilla} \\
\hline
\multirow{3}{*}{JENA}
& FGSM & 1.16 & 9.54   & 9.64   & $8.22\times$  & $8.31\times$  & $1.01\times$ \\
\cline{2-8}
& BIM  & 1.16 & 41.48  & 41.59  & $35.76\times$ & $35.85\times$ & $1.00\times$ \\
\cline{2-8}
& PGD  & 1.16 & 41.45  & 41.62  & $35.73\times$ & $35.88\times$ & $1.00\times$ \\
\hline
\multirow{3}{*}{ECL}
& FGSM & 5.03 & 29.20  & 29.27  & $5.81\times$  & $5.82\times$  & $1.00\times$ \\
\cline{2-8}
& BIM  & 5.03 & 148.43 & 148.85 & $29.51\times$ & $29.59\times$ & $1.00\times$ \\
\cline{2-8}
& PGD  & 5.03 & 148.34 & 149.08 & $29.49\times$ & $29.64\times$ & $1.00\times$ \\
\hline
\multirow{3}{*}{AEP}
& FGSM & 1.75 & 29.49  & 29.55  & $16.85\times$ & $16.89\times$ & $1.00\times$ \\
\cline{2-8}
& BIM  & 1.75 & 159.36 & 159.47 & $91.06\times$ & $91.13\times$ & $1.00\times$ \\
\cline{2-8}
& PGD  & 1.75 & 159.36 & 159.44 & $91.06\times$ & $91.11\times$ & $1.00\times$ \\
\hline
\end{tabular}
\end{table}

MTD is expensive in absolute terms, ranging from $5.81\times$ to $91.13\times$ the base training cost. The overhead is dominated by adversarial fine-tuning of multiple students, which scales with the number of attack iterations and the number of epsilon values in the training sweep. The marginal cost of layer targeting on top of this is negligible: MorphStrata/Vanilla never exceeds $1.01\times$ across all nine conditions. For any deployment that has accepted MTD, MorphStrata incurs negligible marginal cost; the robustness gains on AEP at high perturbation budgets provide the clearest justification.

\section{Memory Footprint}
\label{app:memory}

Peak VRAM and peak CPU RAM are recorded by the monitoring system. All values are in MB. The \textit{Base train} column shows memory during undefended base model training, which is identical for both pipelines. \textit{Vanilla} and \textit{MorphStrata} columns show peak memory during adversarial fine-tuning for each pipeline.

\begin{table}[t]
\centering
\caption{Peak VRAM and CPU RAM during training stages on an NVIDIA L4 GPU (Ada Lovelace). \textit{Base train} is shared across both pipelines.}
\label{tab:memory}
\footnotesize
\renewcommand{\arraystretch}{1.12}
\setlength{\tabcolsep}{2pt}
\begin{tabular}{|c|c|>{\centering\arraybackslash}p{1.5cm}|>{\centering\arraybackslash}p{1.5cm}|>{\centering\arraybackslash}p{1.8cm}|>{\centering\arraybackslash}p{1.5cm}|>{\centering\arraybackslash}p{1.5cm}|>{\centering\arraybackslash}p{1.8cm}|}
\hline
\rowcolor{gray!20}
& &
\multicolumn{3}{c|}{\textbf{VRAM (MB)}} &
\multicolumn{3}{c|}{\textbf{CPU RAM (MB)}} \\
\arrayrulecolor{black}\cline{3-8}
\rowcolor{gray!20}
\multirow{-2}{*}{\textbf{Dataset}} &
\multirow{-2}{*}{\textbf{Attack}} &
\textbf{Base train} &
\textbf{Vanilla} &
\textbf{MorphStrata} &
\textbf{Base train} &
\textbf{Vanilla} &
\textbf{MorphStrata} \\
\hline
\multirow{3}{*}{JENA}
& FGSM & 767 & 977   & 1{,}035 & 1{,}788 & 1{,}822 & 1{,}836 \\
\cline{2-8}
& BIM  & 767 & 977   & 1{,}035 & 1{,}788 & 1{,}822 & 1{,}836 \\
\cline{2-8}
& PGD  & 767 & 977   & 1{,}035 & 1{,}788 & 1{,}823 & 1{,}836 \\
\hline
\multirow{3}{*}{ECL}
& FGSM & 893 & 1{,}285 & 1{,}591 & 1{,}662 & 1{,}744 & 1{,}861 \\
\cline{2-8}
& BIM  & 893 & 1{,}285 & 1{,}593 & 1{,}662 & 1{,}746 & 1{,}864 \\
\cline{2-8}
& PGD  & 893 & 1{,}285 & 1{,}593 & 1{,}662 & 1{,}747 & 1{,}865 \\
\hline
\multirow{3}{*}{AEP}
& FGSM & 1{,}277 & 2{,}871 & 3{,}279 & 7{,}583 & 7{,}781 & 7{,}863 \\
\cline{2-8}
& BIM  & 1{,}277 & 2{,}871 & 3{,}279 & 7{,}583 & 7{,}789 & 7{,}859 \\
\cline{2-8}
& PGD  & 1{,}277 & 2{,}859 & 3{,}257 & 7{,}583 & 7{,}788 & 7{,}865 \\
\hline
\end{tabular}
\end{table}

MorphStrata requires modestly more VRAM than Vanilla during adversarial training, peaking at a 24\% increase for ECL (1{,}285\,MB to 1{,}591\,MB). CPU RAM increases are under 7\% across all conditions. Adversarial training consumes roughly $2\times$ to $3\times$ the VRAM of clean base training because the attack loop holds perturbation tensors and gradients in memory simultaneously. AEP CPU RAM is substantially higher than JENA or ECL ($\approx$7.5\,GB vs.\ 1.8\,GB) because the AEP notebook retains the full preprocessed dataset in CPU memory throughout training.

\section{Experimental Platform}
\label{app:platform}

All experiments were conducted using the hardware summarized in Table~\ref{tab:platform}. The GPU is an NVIDIA L4 (Ada Lovelace architecture, compute capability 8.9), Its 22.5\,GB of VRAM was sufficient to hold all student models and their adversarial perturbation tensors in memory simultaneously, with no gradient checkpointing required. The 12-thread Intel Xeon CPU and 52\,GB of system RAM meant that even the AEP pipeline, which retains its full preprocessed dataset in CPU memory, was not memory-constrained. Framework versions are fixed at PyTorch 2.10.0 with CUDA 12.8; all reported results are reproducible under these versions.

\begin{table}[h]
\centering
\caption{Experimental platform specifications.}
\label{tab:platform}
\footnotesize
\renewcommand{\arraystretch}{1.12}
\setlength{\tabcolsep}{2pt}
\begin{tabular}{|>{\raggedright\arraybackslash}p{3.8cm}|>{\raggedright\arraybackslash}p{6.0cm}|}
\hline
\rowcolor{gray!20}
\textbf{Component} & \textbf{Specification} \\
\hline
GPU                & NVIDIA L4 (Ada Lovelace, compute cap.\ 8.9) \\
\hline
GPU VRAM           & 22.5\,GB (23{,}034\,MiB) \\
\hline
CPU                & Intel Xeon @ 2.20\,GHz \\
\hline
CPU cores / threads & 6 cores, 12 threads (2 threads/core) \\
\hline
System RAM         & 52\,GB \\
\hline
\end{tabular}
\end{table}

\end{document}